\documentclass[letterpaper, 10pt, conference]{ieeeconf}

\IEEEoverridecommandlockouts

\makeatletter
\let\NAT@parse\undefined
\makeatother

\usepackage{graphics} 
\usepackage{graphicx}
\usepackage{epsfig} 
\usepackage{mathptmx} 
\usepackage{times} 
\usepackage{amsmath} 
\usepackage{amssymb}  
\usepackage{subfigure}
\usepackage{balance} 
\usepackage[]{caption2}

\usepackage{float}
\usepackage{gensymb}
\usepackage{textcomp}
\usepackage{flushend} 

\title{\LARGE \bf
Learning Autonomous Mobility Using Real Demonstration Data 
}

\author{Jiacheng Gu and Zhibin Li
\thanks{Jiacheng Gu and Zhibin Li are with the Institute for Perception, Action, and Behaviour, School of Informatics, University of Edinburgh, 10 Crichton St, Edinburgh EH8 9AB, United Kingdom.
        \newline Email: {\tt\small J.Gu@ed.ac.uk}}%
}

\begin{document}

\maketitle
\thispagestyle{empty}
\pagestyle{empty}

\begin{abstract}

This work proposed an efficient learning-based framework to learn feedback control policies from human teleoperated demonstrations, which achieved obstacle negotiation, staircase traversal, slipping control and parcel delivery for a tracked robot. 
Due to uncertainties in real-world scenarios, eg obstacle and slippage, closed-loop feedback control plays an important role in improving robustness and resilience, but the control laws are difficult to program manually for achieving autonomous behaviours. We formulated an architecture based on a long-short-term-memory (LSTM) neural network, which effectively learn reactive control policies from human demonstrations. Using datasets from a few real demonstrations, our algorithm can directly learn successful policies, including obstacle-negotiation, stair-climbing and delivery, fall recovery and corrective control of slippage. We proposed decomposition of complex robot actions to reduce the difficulty of learning the long-term dependencies. Furthermore, we proposed a method to efficiently handle non-optimal demos and to learn new skills, since collecting enough demonstration can be time-consuming and sometimes very difficult on a real robotic system.

\end{abstract}

\section{Introduction}

\begin{figure}[t]
    \centering
    \includegraphics[width=0.5\textwidth]{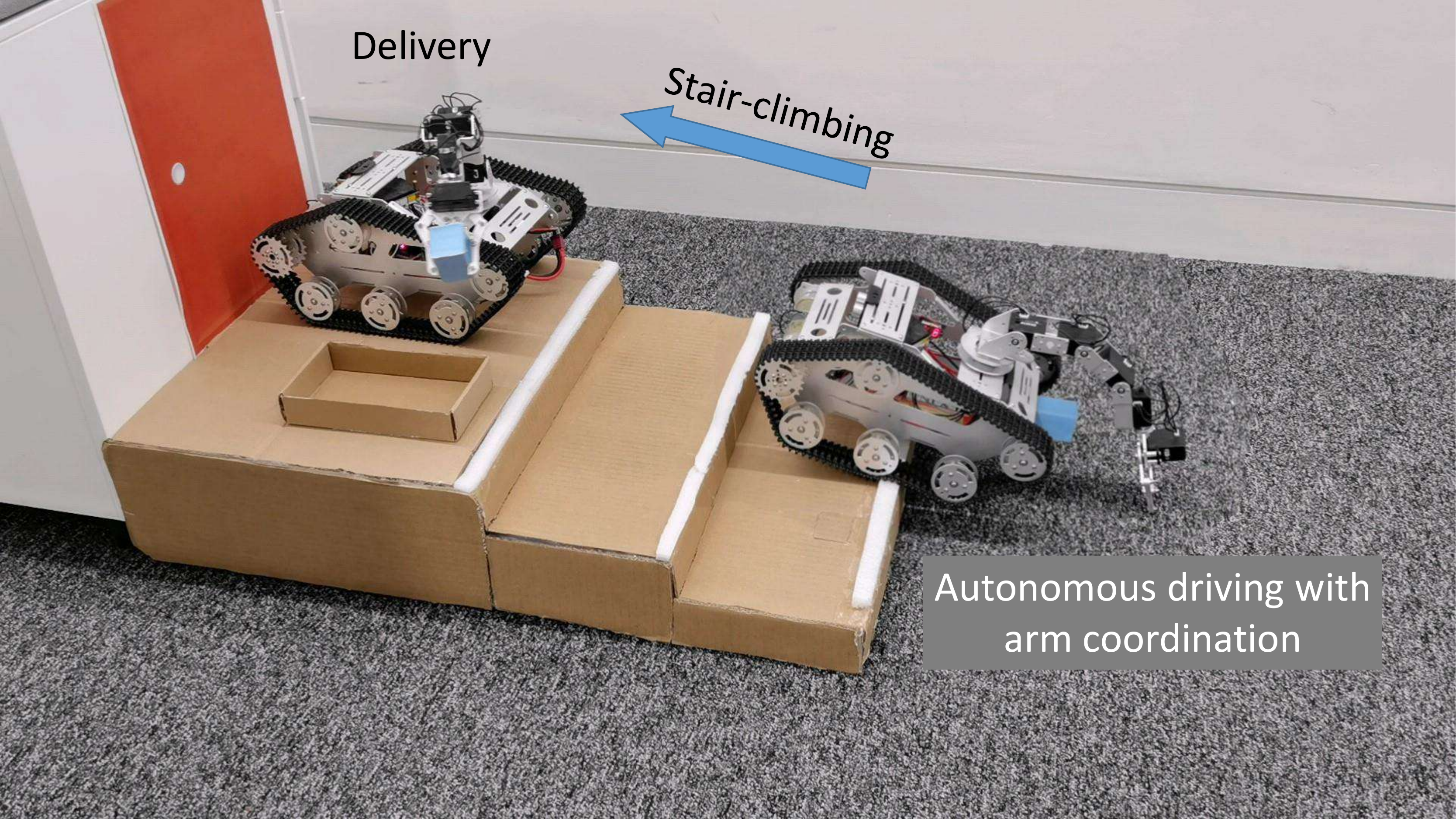}
    \caption{Adaptive feedback control using learning from demonstration for autonomous mobility and manipulation.} 
    \label{fig:intro}
\end{figure}

Nowadays mobile robots are able to move on most of the even roads and handle situations using perception, localisation and path planning, but still have difficulties on complex, slippery and uneven terrains with obstacles and stairs, for example, as shown in the autonomous delivery scenarios (Fig.\ref{fig:intro}). However, slippage and external disturbance usually results in mobility failures, which can result in unacceptable damages for both the robot and the surroundings. Therefore, robust and resilient controls are needed to properly react to slippage and disturbance in complex scenarios, in order to achieve autonomous behaviors.

Deep reinforcement learning (DRL) is powerful to acquire a wide range of skills \cite{yang2018learning}, especially while training in simulations. However, for mobile robots, especially the tracked ones, it is still very difficult for the state-of-art physics engines to simulate the material properties of tracks and their interactions with the environment, when there are significant slippage and deformation. Meanwhile, the data expensive nature make DRL approach extremely expensive and unsafe to explore on a physical robot. Therefore, we are interested in learning and transferring a policy directly from a complex hardware system and real data, which are not possible to be simulated in physics simulations.

Learning from Demonstration (LfD) is a method that enable the agent to acquire skills by learning from the expert's demonstrations. It has been widely used in robotics area. e.g. autonomous driving \cite{mycite:carbehaviourclone}, \cite{mycite:car1}, \cite{mycite:car2}, flight \cite{mycite:flight}, manipulation \cite{mycite:manipulation1}, \cite{mycite:manipulation2} and so on. In order to better learn the expert's policy, most LfD algorithms require a large amount of high-quality demonstrations.

There are mainly two classes of LfD, Inverse Reinforcement Learning (IRL) and behavioural cloning. IRL \cite{mycite:inverseRL} requires a reward function that able to describe the demonstration as (or similar to) the optimal behaviour. Behavioural cloning \cite{mycite:behaviouralCloning} is a supervised learning algorithm that optimises a neural network mapping from observations to actions. For IRL, exploration in real-world is likely to cause physical damage to the robot. While even a good policy is learnt in a simulation environment, it does not usually perform well in reality because of sim2real problem. We therefore use the behavioural cloning algorithm in our research.

Many successful applications use behavioural cloning algorithm in robotics, especially when the dimension of environment and action states is low. For example, \cite{mycite:lowDimensionBC3} extracts spatial information and object-recognition information from visual input to learn robotics motor skills. Sometimes it is challenging to extract useful information from multiple sensors. Therefore many researchers use raw image or video as the demonstration to train the neural network \cite{mycite:BCfromRawimg}, \cite{mycite:drones}. However, we believe that human always receive many redundant information and human brain can extract few useful information. So simply using raw image data would increase the complexity of the task. Our controller only requires few necessary information and can generalise well.

Generalisation on unseen states is challenging for behavioural cloning. Dataset Aggregation \cite{mycite:behaviouralCloning} requires multiple iterations to train on aggregated data. Some researchers \cite{mycite:useSimulation} use demo in simulation to replace real demonstration. However, our model can not be simulated and they both has sim2real problems. Some other researchers use cloud-based data collection techniques. \cite{mycite:cloud1} uses the Google object recognition engine, \cite{mycite:cloud2} presents their software tool that enables crowd-sourcing, so is \cite{mycite:cloud3}.
We proposed a method to cut down the non-optimal part of the demonstration and merge a short, near-optimal demo with the original one. Then the controller could learn the policy better or learn new skills. This method could also handle non-optimal demos.

Once have a new near-optimal demo, re-training the network from scratch (on a better demo) is time-consuming and unnecessary. We use transfer learning (TL) to greatly speed up this process. TL focuses on storing knowledge gained while solving one problem (source domain) and applying it to a different but related problem (target domain). Some robotics researchers \cite{mycite:tf_sim2real1}, \cite{mycite:tf_sim2real2} use transfer learning to solve sim2real problems. In our experiment, the control policy is learnt from original demonstrations (the source domain), then we transfer it to the new near-optimal demo (the target domain) to learn new policy.

In this paper, we proposed an LSTM-based neural network controller capable of learning reactive and responsive policies for \textit{autonomous driving control} from a few demonstrations from human teleoperators. LSTM has been used for robot control, such as in \cite{mycite:lstm1}, where LSTM was used as temporal encoders for information processed by CNN from side-view cameras, only the current data was fed to LSTM network as LSTM has `memory cells' to maintain long-term memory. However, our experiments shows that an implicit use of the memory mechanism of LSTM has limited performance. Instead, an explicit use of several past datapoints and the current one can significantly improve the autonomous performance. 

Main contributions of this paper are summarised below:
\begin{itemize}
    \item We formulate the use of a set of sequence feedback states containing history data for a LSTM based robot controller, which yeilds better performance than only using the current data; 
    \item We decompose the autonomous delivery task and develop mobility and manipulation controllers of the robot, reducing the difficulty of long-term dependency in LSTM.
    \item We apply transfer learning in LfD problems to efficiently learn new skills and solve problems caused by non-optimal demonstrations.
\end{itemize}

The paper is organised as follows. Section \ref{sec:robot system} introduces the proposed robot system in detail, and presents the control architecture of the neural network controller. In Section \ref{sec:experiments}, we describe our setup and analyse the experimental results. Conclusion and future work are presented in Section \ref{sec:conclusion}.

\section{Robot System}
\label{sec:robot system}

\subsection{Hardware Architecture}
\label{sec:systm-architecture}

For the proof of concept, we validate the learning algorithms using a tracked robot as shown in Fig.\ref{fig:intro}, which has a 6-degree-of-freedom (DoF) robot arm on the top and two DC geared motors to directly drive each track. This also allows us to set up mockup testing environment more easily, for example, we can have a scaled up size of stairs that is quite large compared to the size of the robot, such that the tracked robot needs to learn \textit{autonomous whole-body coordination of using the arm to go up stairs and recover from tipping over}. These are the advantages of a miniaturised setup that the normal real hardware cannot afford to risk, so we can push the testing of robustness and performance to the extreme. 

We use \emph{AT Mega328 UNO} as the Micro-controller Unit (MCU), it controls the DC geared motors and the arm on the robot. MCU collects data from on-board sensors (including IMU \emph{MPU6050} and sonic sensor \emph{HC-SR04}), and sends these data to the computer via Bluetooth module \emph{HC-05}.
A PS4 joystick is connected to the computer to collect the human operator's control command. Another Bluetooth module \emph{HC-05} is connected to the computer as well for data communication with our robot.

\subsection{Control Modules}
Our approach requires proper demos executed by an expert, from which the neural network controller could learn the proper control policy for the robot to go through a complex area (with different shapes of obstacles and stairs). The learnt policy needs to be robust enough to handle the potential slippage and recover from failures. In each demo, the tracked robot is teleoperated to go over several obstacles and climb the stair model, the robot stops to delivery the package when the sonic sensor detects a door (height \textgreater 25cm) in front. Then it goes backwards.

\subsubsection{Arm Control}
Once the low-level robot arm controller receives joints' position $p$ and moving time $t$ from the MCU, each joint would start moving to the target position $p_i$ in time $t$. (New coming command would overwrite the previous one if the previous action has not finished). 

We design 5 different motion primitives for the robot arm in the task, including move or keep the arm in the middle (idle) position, sustain the robot in the back (for stair-climbing or going over obstacles), failure-recovery from left or right and delivery. One channel is used to  represent 5 different arm actions (states): middle (0), back (1), failure-recovery from left (2), failure-recovery from right (3) and delivery (4), arm control actions $u_a\in \{0,1,2,3,4\}$.

\subsubsection{Moving Control}
In this paper, we focused on training a low-level motion controller that can react to real-time sensor input and naively going forward until delivering the package, fixing control errors or failures. In this task, the robot ought to keep going forward while the front area is clean, go over any possible obstacles or stairs, delivering the package while facing a wall and go back once finished. 

As the gear ratio of the motor's gearbox is large (100:1), so we use the maximum motor speed for the moving action. The control action includes: go backwards (-1), stop (0) and go forward (1). $v_l$ and $v_r$ are left and right motor speed respectively, then:
\begin{align}
    v_l = v_r = u_m,
\label{eq:moving}
\end{align}
where $u_m \in \{-1,0,1\}$ is the control signal for moving forward or backwards.
The steering control corrects the robot's heading direction after failure-recovery, going over an obstacle or external disturbance. We use one channel to present 3 different steering states: turning left, turning right and idle (not turning). While turning left, the robot's left motor would go backwards at full speed and its right motor would go forward and vice versa. While the expert presses going forward and turning left at the same time, the left motor of the robot would stop and the right one would go forward at full speed. The steering control actions are: turn left (-1), idle (0), turn right (1). $u_s \in \{-1,0,1\}$ is the steering control actions, the velocity of left and right motor gives:

\begin{equation}
\left\{
     \begin{array}{lr}
     v_l = max\{min\{u_m + u_s, 1\}, -1\}, &  \\
     v_r = max\{min\{u_m - u_s, 1\}, -1\}, &  
     \end{array}
\right.
\end{equation}
therefore, motor velocity $v_l, v_r \in \{-1,0,1\}$.

We use 3 channels to control arm motion (5 actions), movement (3 actions) and steering (3 actions). There are totally 45 possible action combinations and 10 of them are used in this task (Table \ref{tab:ActionDef}). We encode the aforementioned action combinations to 45 motion primitives for the network to classify:
\begin{align}
    ID_a = u_a \times 9 + (u_s + 1) \times 3 + u_m + 1.
\label{eq:actionIndex}
\end{align}

\begin{figure}[t]  
    \centering
    \includegraphics[width=0.5\textwidth]{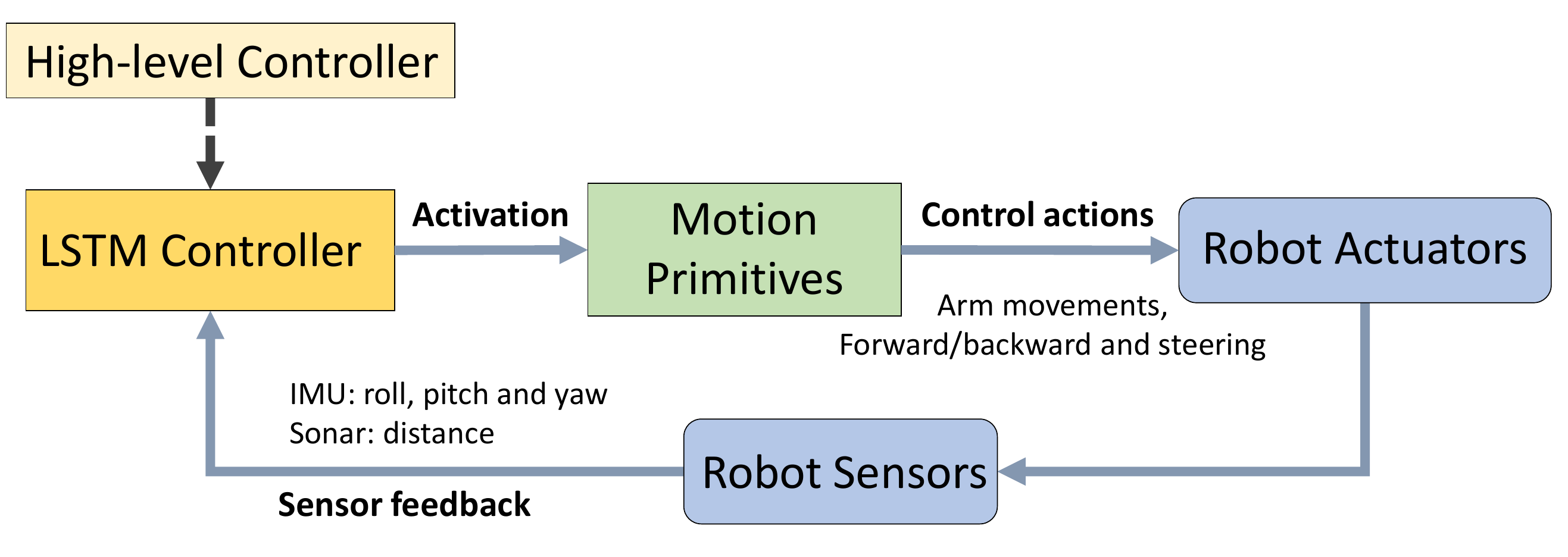}
    \caption{System control diagram.}
    \label{fig:controlDiagram}
\end{figure}

\begin{table}[t]
\caption{Definition of The Used Actions}\label{tab:ActionDef}
    \begin{center}
    \begin{tabular}{cccc}
    \hline
    Action Index ($ID_a$) & Robot Arm & Steering & Movement \\
    \hline
    1  & Middle   & Left  & Stop \\
    3  & Middle   & Idle  & Backward \\
    4  & Middle   & Idle  & Stop \\
    5  & Middle   & Idle  & Forward \\
    7  & Middle   & Right & Stop \\
    13 & Back     & Idle  & Stop \\
    14 & Back     & Idle  & Forward \\
    22 & FR left  & Idle  & Stop \\
    31 & FR right & Idle  & Stop \\
    40 & Delivery & Idle  & Stop \\
    \hline
    \end{tabular}
    \end{center}
\end{table}

\begin{figure}[t]
    \centering
    \includegraphics[width=0.5\textwidth]{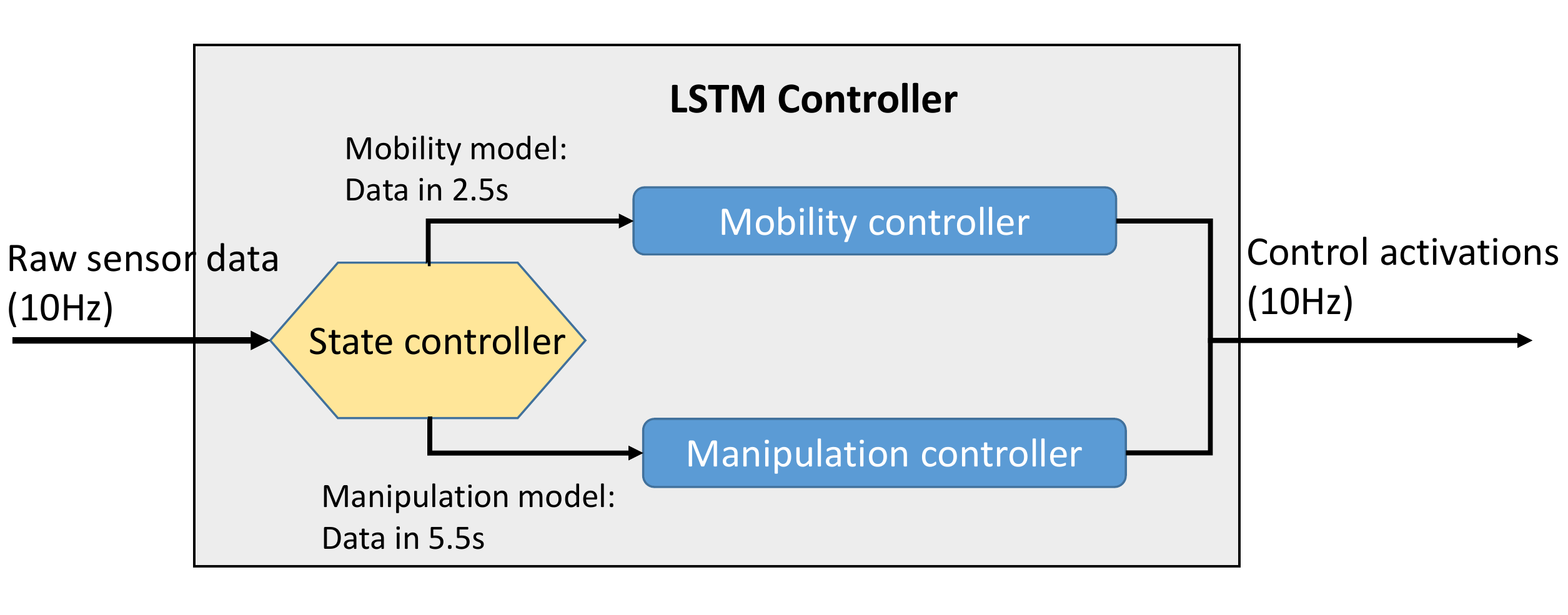}
    \caption{LSTM controllers for mobility and manipulation.}
    \label{fig:2lstm}
\end{figure}
\subsection{Control}
The aforementioned robot system requires a low-level reactive responsive controller to handle different situations while moving. Based on high-level control commands and robot's sensor feedback, the proposed LSTM controller is capable of generating proper control actions.

The control diagram is shown in Fig.\ref{fig:controlDiagram}. 
In our task, commands from the high-level control is: keep moving forwards until facing a wall, then deliver the package and go back. The low-level reactive controller minimises yaw differences (compare to reference orientation) by steering control and keeps the robot stable (failure-recovery). The high-level controller can be replaced by any path planner or an intermittent, high-delay remote control signal. 
We decompose the task into mobility and manipulation. Two separate controllers control each part. A state machine (\ref{fig:2lstm}  robot state controller) controls the transition between two working models. The robot is in mobility model after it starts. It turns into manipulation model when facing a wall, meanwhile, a manipulation timer is started. Finally, it turns back to mobility model after the manipulation action is finished.

In generalisation, the LSTM controller outputs digital numbers to activate motion primitives, which control the actuators on the robot. The on-board sensors on the robot would send real-time sensor measurements (IMU and sonar) to the controller.

\begin{figure}[t]
    \centering
    \includegraphics[width=0.5\textwidth]{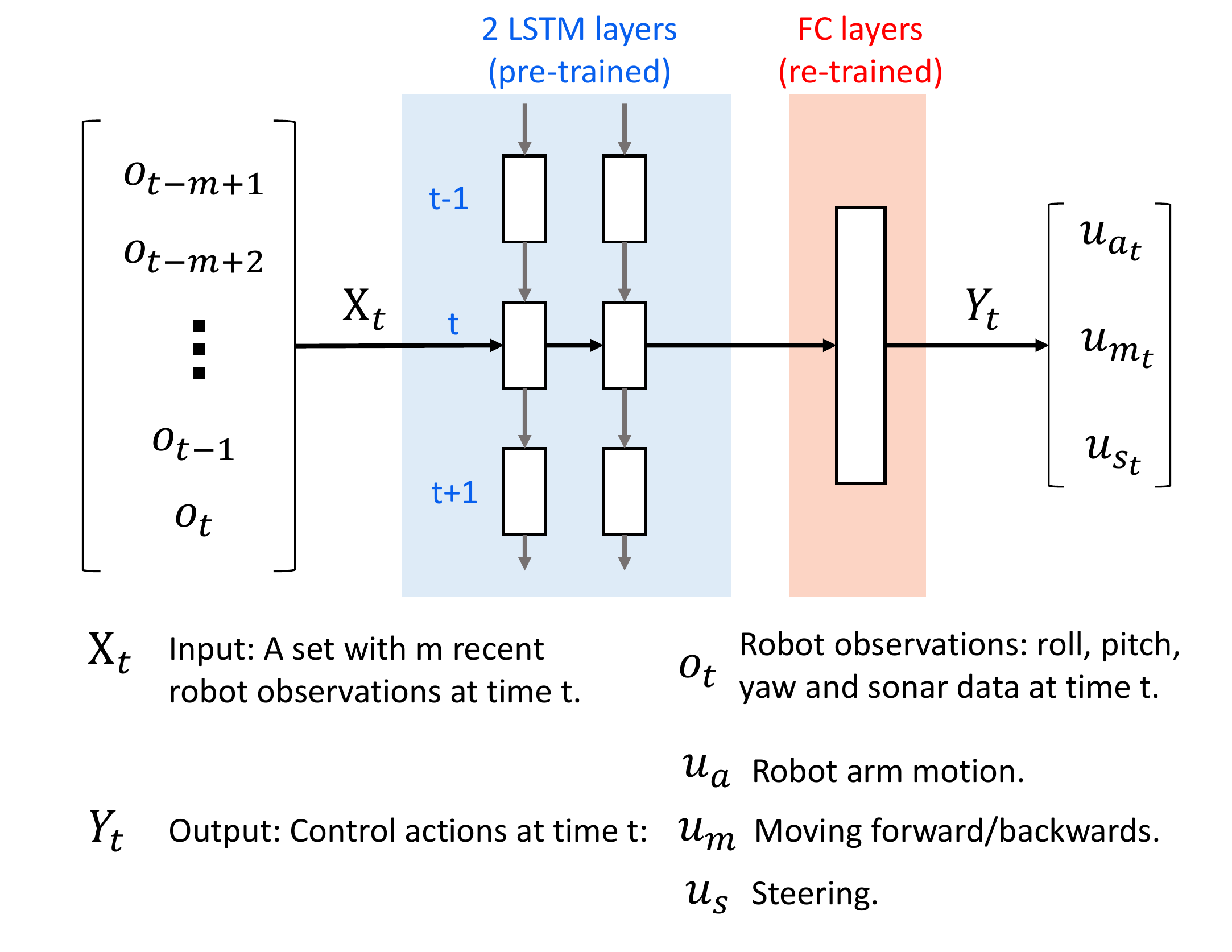}
    \caption{The proposed architecture based on LSTM.}
    \label{fig:lstm}
\end{figure}

\subsection{Neural Network Controllers}
To solve the multi-class classification problem in our task, we propose a control framework based on behavioural cloning algorithm to learn the human operator's control policy while teleoperating the tracked robot from few proper demonstrations. We collect a dataset $ D_{task}={(o_t,u_t)^{(i)}}$ composed of pairs of observation $o_t$ and corresponding control action $u_t$ at time $t$.  Where $i$ indicates the index of multiple demonstrations. Given the observation and control, the two neural networks with inner parameters $\theta$ optimises the control policy $\pi_\theta(u_t|o_t)$, which is a function that maps from input observations (and several sets of previous data) to output control actions.

More specifically, the neural networks classify the outputs into 10 different possible motion primitives based on the current and previous inputs. The observation (input) $o_t=[y_t,p_t,r_t,d_t]$ consists of the yaw $y_t$, pitch $p_t$ and roll $r_t$ of the robot and distance $d_t$ in the front. The control (output) consists of robot arm control action $u_a\in \{0,1,2,3,4\}$, moving control action $u_m \in \{-1,0,1\}$ and steering control action $u_s \in \{-1,0,1\}$. We use one-hot encoding for the neural network output.

The sampling frequency of IMU is 30 Hz. Then a first-order Butterworth low-pass filter with 15 Hz cut-off frequency is applied for denoising and removing outliers. The sampling frequency of the sonic sensor is 20 Hz. According to \cite{mycite:humanreactiontime}, human's average reaction time is around 250 ms (4 Hz). A top computer game player can react to an incident in 100 ms. In order to mimic a human's reaction, we set our network's control frequency at 10 Hz. (Note that human's decision-making process is much faster, the aforementioned reaction speed is limited by human body.)


We use two 2-layer LSTM neural networks as the mobility controller and manipulation controller (Fig.\ref{fig:lstm}). Given the current observation $o_t$ and $m-1$ previous observations, the trained network could generate the control action $Y_{t}$ ($t\geq m$), which should be able to control the tracked robot to climb the stair properly, similar to human operator's control in demonstrations. Concretely, we pack up $m$ recent observations:

\begin{align}
    X_{t}=\{o_{t-m+1},o_{t-m},...,o_t\}, t\geq m,
\label{eq:X}
\end{align}
and feed them into the LSTM. The LSTM with optimised the policy $\pi_\theta(u_t|o_t)$ then outputs the control action:
\begin{align}
    Y_{t}=LSTM(X_{t};\theta), t\geq m,
\label{eq:y}
\end{align}
where m is the timestep of the LSTM. 25 timesteps is chosen for the mobility controller, which represents a 2.5s time-window in real world (10 Hz control frequency). It can cover the longest mobile action (2.1s plus idle time that during actions) in the task. We choose 5.5 time window for the manipulation controller. Table \ref{tab:duration} shows the duration of some actions in the demonstration. 

\begin{table}[t]
\caption{Duration for some actions}\label{tab:duration}
\begin{center}
\begin{tabular}{ccc}
\hline
Action  & Duration [s] & Description \\
\hline
Arm move to back            & 1.0           &  ---\\
Arm move to middle          & 1.0           &  ---\\
Arm failure-recovery        & 2.0           &  ---\\
Go forward, arm back        & (around) 1.2  &  Go over obstacles\\
Go forward, arm back        & (around) 1.5  &  Climb stairs\\
Idle (human)                &  0.2-0.4      & Reaction time  \\
Steering                    & 0.5-2.1       & Yaw diff.: $20^\circ-90^\circ$ \\
Delivery                    & 5.0           &  ---\\

\hline
\end{tabular}
\end{center}
\end{table}

The proposed framework use softmax cross-entropy loss:
\begin{align}
    \sigma(z_j) = softmax(z_j)=\frac{e^{z_j}}{\sum_{c=0}^{C-1}e^{z_c}},
\label{eq:softmax}
\end{align}

For this classification problem, 45 actions are not evenly distributed, so each term of cross-entropy is scaled with the corresponding weight.
\begin{align}
    loss = -\sum_{c=0}^{C-1}w_c l_c \ln{\sigma(z_c)},
\label{eq:weightedSoftmaxEntropy}
\end{align}
where $l$ is the one-hot labels ($\sum l=1$), $z$ is the output of the fully-connected cell, it has the same shape as $l$ ($1 \times 4536$ in this task). $C$ is motion primitives ($C=45$ in this task), and $w_c$ is the weight of 45 different motion primitives in each dataset. We use Adam optimiser to train our LSTM, the learning rate chosen for this task is $0.001$. There are 800 hidden units in each LSTM layer of two controllers.

\section{Experiments}
\label{sec:experiments}

\begin{figure}[t]
    \centering
    \includegraphics[width=0.5\textwidth]{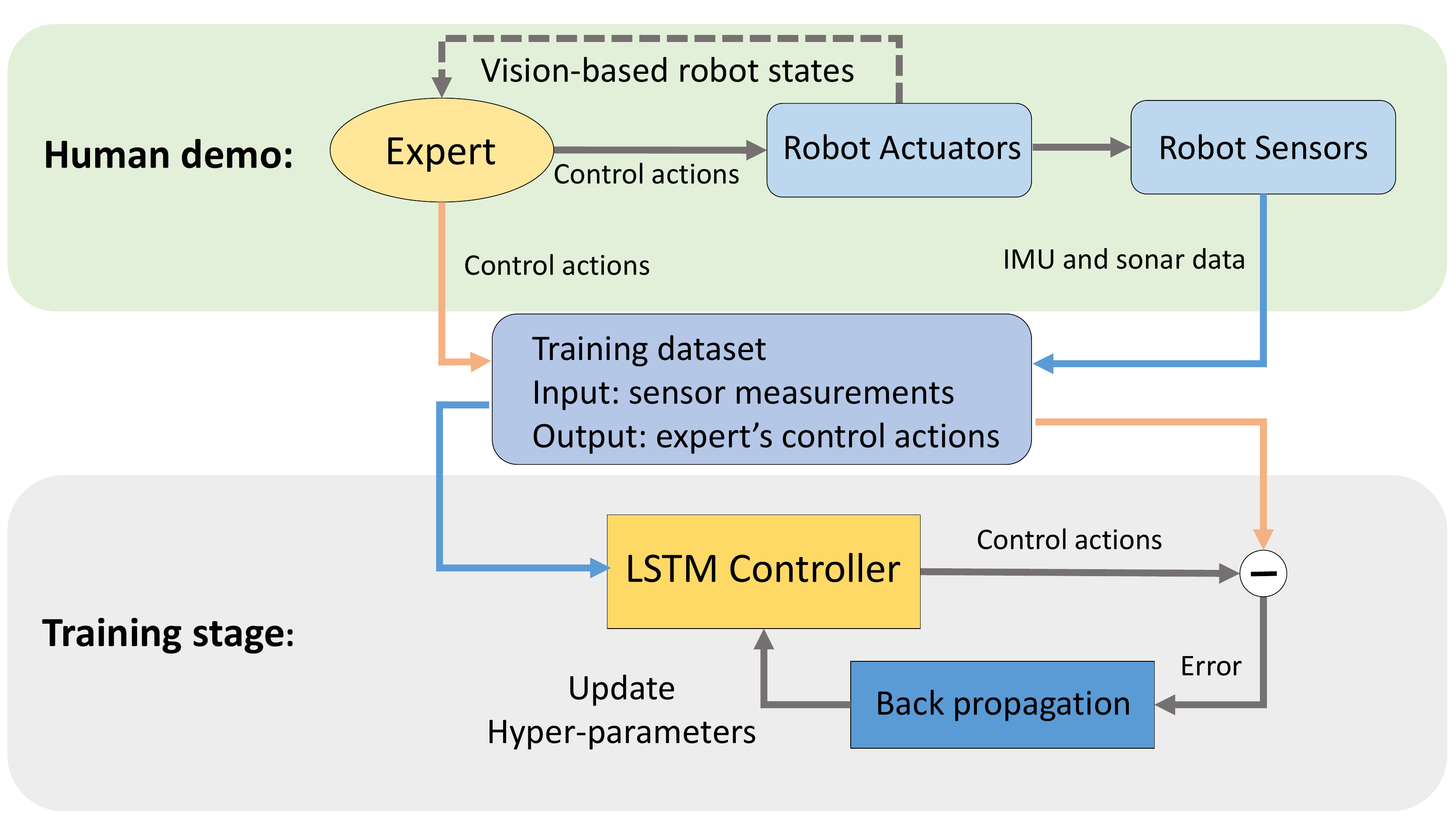}
    \caption{Training procedure.}
    \label{fig:training}
\end{figure}

We collect a set of demonstration data to train the proposed network, it achieves human-level reactive control in experimental environment. (Fig.\ref{fig:training} shows the collecting and training procedure.) Then we use transfer learning to efficiently learn new skills for a better demo.

When using a single LSTM to control both mobility and manipulation, the time-window has to be large enough to cover the longest action: delivery (5s). In this case, the network would be less sensitive to small variation of sensor data and might fail to react to it. Computing cost would also be much higher. If a small time window is chosen, the controller is likely to get stuck after facing a wall. As the long-term dependency of LSTM is hard to keep (transition of sonar data and delivery action takes at least 10s, which gives more than 100 previous robot states).

So we train the mobility controller and the manipulation controller separately. A finite state machine is use to control the working models of the robot (Fig.\ref{fig:2lstm}). 

\begin{figure*}[t]
\centering
    \subfigure[]{
    \begin{minipage}[t]{0.14\linewidth}
    \centering
    \includegraphics[width=1\linewidth]{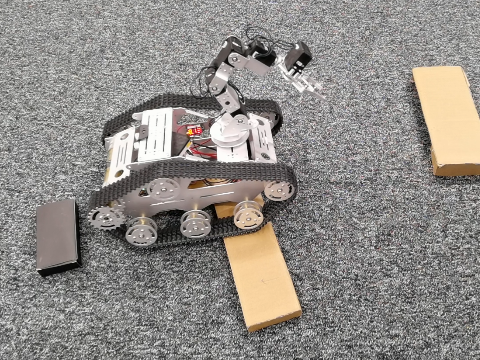}
    \end{minipage}%
    }
    \subfigure[]{
    \begin{minipage}[t]{0.14\linewidth}
    \centering
    \includegraphics[width=1\linewidth]{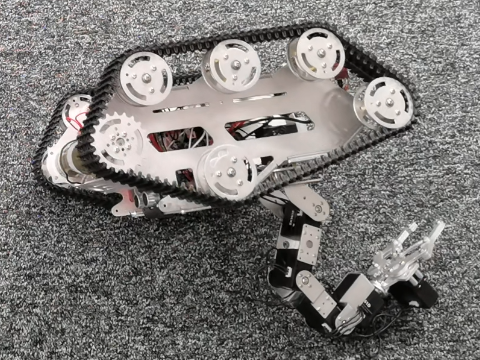}
    \end{minipage}
    }
    \subfigure[]{
    \begin{minipage}[t]{0.14\linewidth}
    \centering
    \includegraphics[width=1\linewidth]{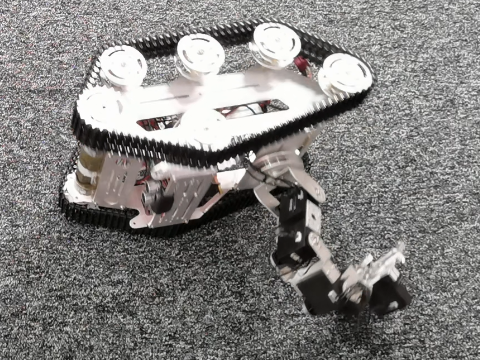}
    \end{minipage}
    }
    \subfigure[]{
    \begin{minipage}[t]{0.14\linewidth}
    \centering
    \includegraphics[width=1\linewidth]{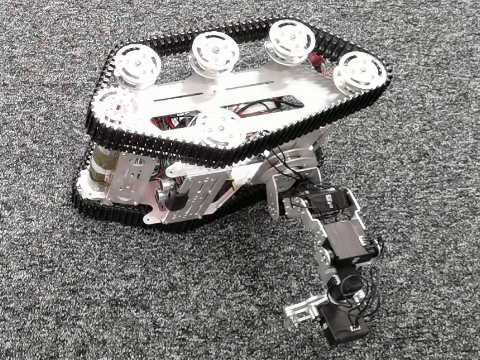}
    \end{minipage}
    }
    \subfigure[]{
    \begin{minipage}[t]{0.14\linewidth}
    \centering
    \includegraphics[width=1\linewidth]{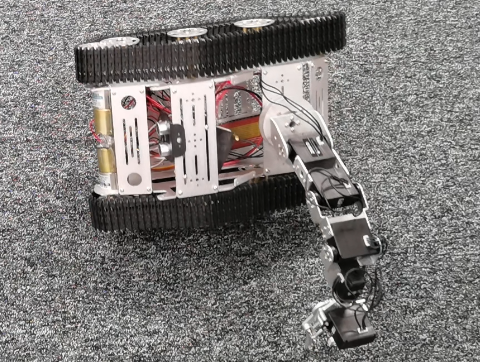}
    \end{minipage}
    }
    \subfigure[]{
    \begin{minipage}[t]{0.14\linewidth}
    \centering
    \includegraphics[width=1\linewidth]{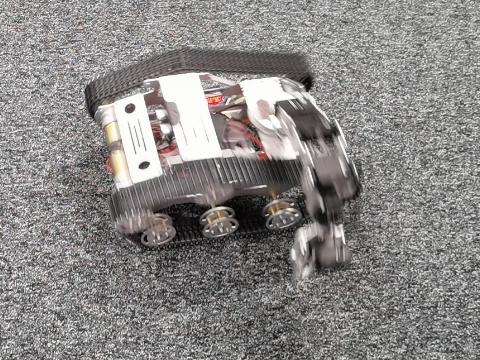}
    \end{minipage}
    }
    \subfigure[]{
    \begin{minipage}[t]{0.14\linewidth}
    \centering
    \includegraphics[width=1\linewidth]{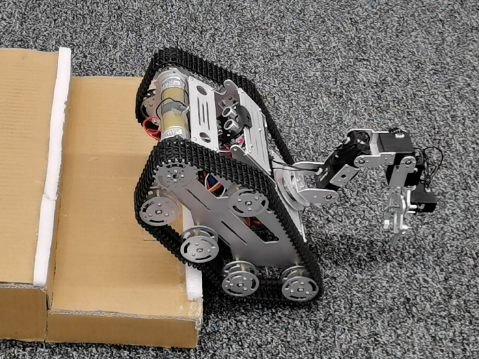}
    \end{minipage}
    }
    \subfigure[]{
    \begin{minipage}[t]{0.14\linewidth}
    \centering
    \includegraphics[width=1\linewidth]{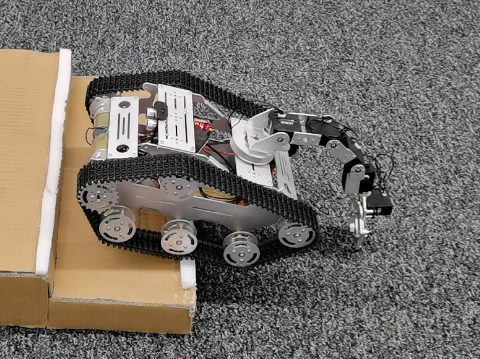}
    \end{minipage}
    }
    \subfigure[]{
    \begin{minipage}[t]{0.14\linewidth}
    \centering
    \includegraphics[width=1\linewidth]{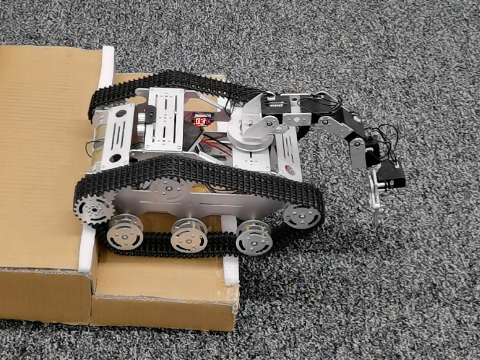}
    \end{minipage}
    }
    \subfigure[]{
    \begin{minipage}[t]{0.14\linewidth}
    \centering
    \includegraphics[width=1\linewidth]{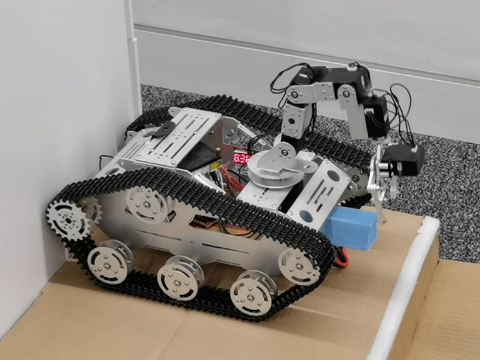}
    \end{minipage}
    }
    \subfigure[]{
    \begin{minipage}[t]{0.14\linewidth}
    \centering
    \includegraphics[width=1\linewidth]{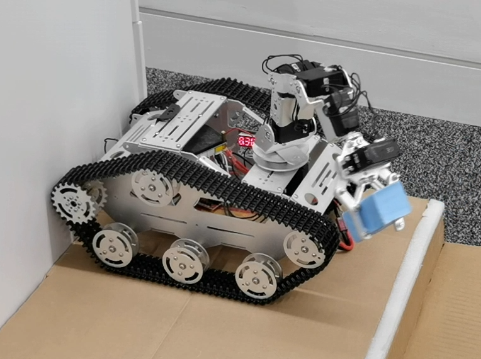}
    \end{minipage}
    }
    \subfigure[]{
    \begin{minipage}[t]{0.14\linewidth}
    \centering
    \includegraphics[width=1\linewidth]{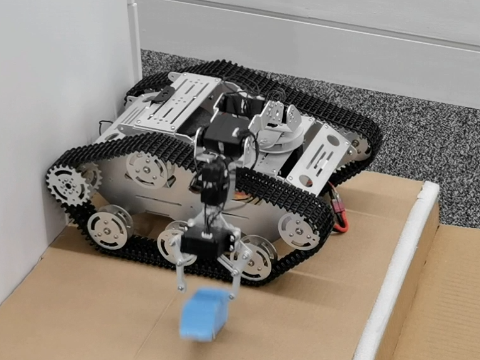}
    \end{minipage}
    }

    \caption{Expert demonstrations: obstacle-negotiation (1), failure-recovery (2-6), stair-climbing(1-3), delivery (10-12).}
    \label{fig:demosnapshot}
\end{figure*}

\begin{figure}[t]
    \centering
    \includegraphics[width=0.5\textwidth]{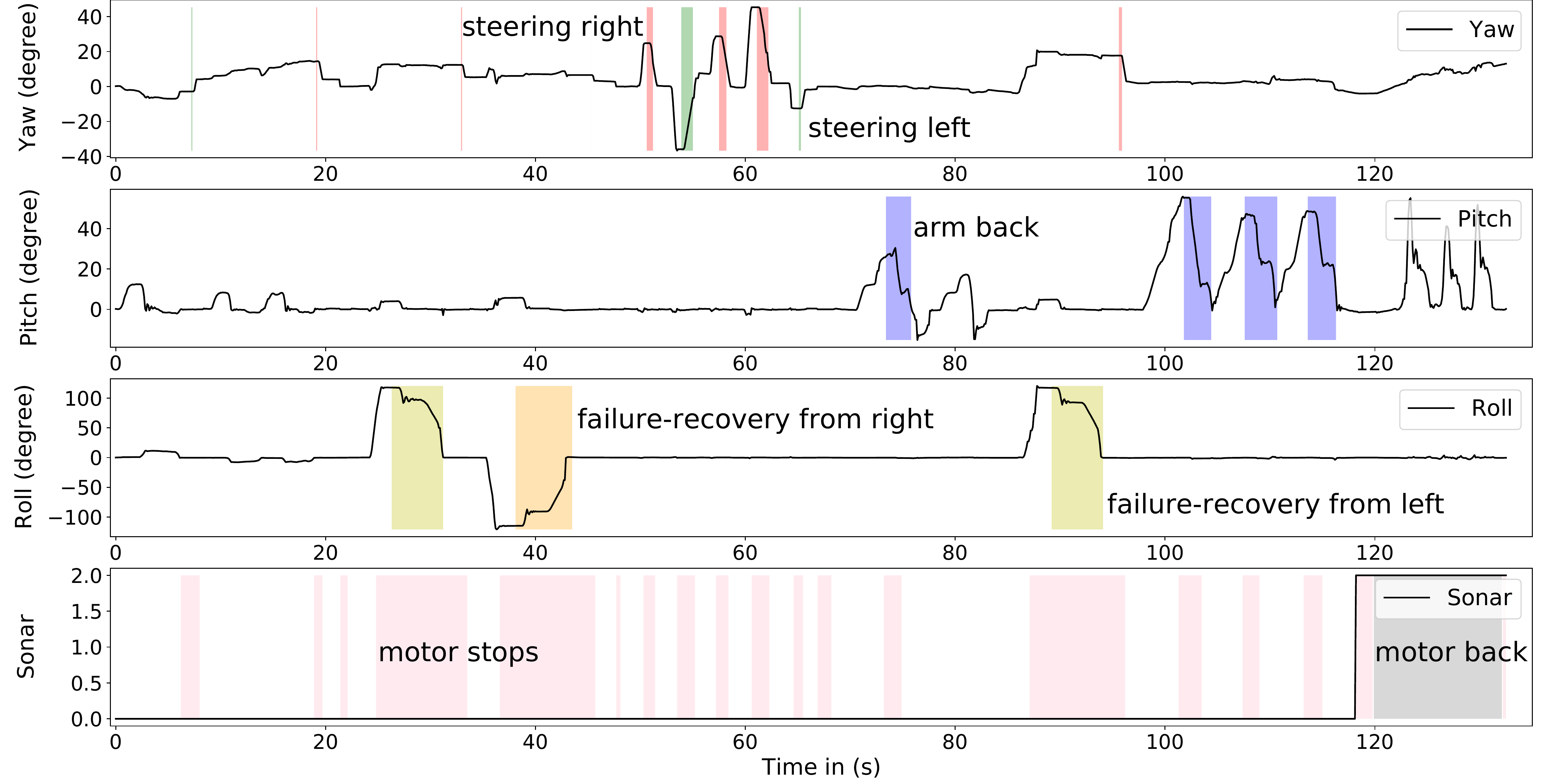}
    \caption{Sensor observations and human operator's control actions (mobility controller).}
    \label{fig:demo1}
\end{figure}

\subsection{Demonstrations and Training}
\label{subsec:training}
The mobility controller enable the robot to go forward (until it faces a wall or door) and go back. It has the capability of obstacle-negotiation, failure-recovery, self-steering and stair-climbing. Once the robot is facing a wall (or door), the state machine would suspend the mobility controller and activate the manipulation controller, then the robot would deliver the package. After the delivery action is finished, the manipulation controller would be deactivated and the mobility controller would be activated again.

\subsubsection{The mobility controller}
In order to train the reactive responsive mobility controller, we collect the following demonstration: In the beginning, the robot goes over a large obstacle (with the help of robot arm) and a small obstacle. Then it recovers from failure in the left and steers right to correct the heading direction after being pushed over. Finally, the robot climbs the stairs and goes back after facing the wall(door). 

This first demo covers most skills needed for the robot in our task, but apparently it could not generalise well for a new unseen scenario (e.g. failure-recovery from right, steers from different yaw angles, driving over obstacle of different size and shapes). Meanwhile, collecting a perfect demonstration that long enough for the robot to learn all the necessary skills is difficult, especially when the task is complex.
Especially, when collecting demos for steering control (the robot was perturbed to change the heading angle, then the expert steers to correct the yaw), the expert is likely to steer too much that the network would fail to learn precise turning time (the turned angle is depend on this turning time).

We solve this problem by trimming the non-optimal parts and merging the existing demo with several short and near-optimal demonstrations. New skills could also be easily learnt by merging a short demo containing these new skills with the original one.
Note that while cutting down the non-optimal part of demonstrations, the robot states would be `deleted'. As the proposed robot system is stable in the `idle state' (robot arm in the idle position, no steering or movement), so the starting state and the ending state of the cut-down demonstration need to be the `idle state'. In this case the missing states do not effect the robot dynamics.  

\begin{figure}[t]
    \centering
    \includegraphics[width=0.5\textwidth]{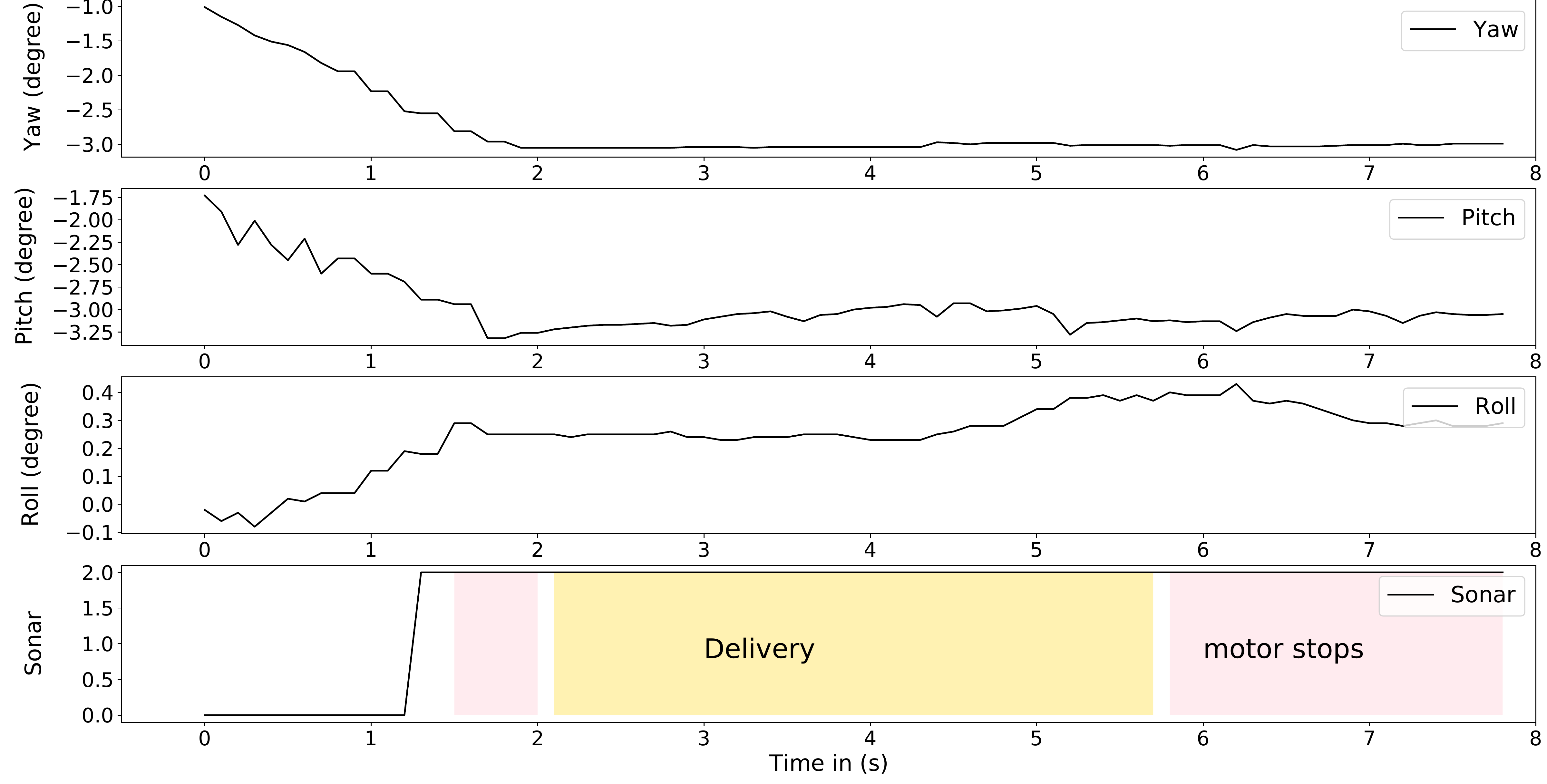}
    \caption{Sensor observations and human operator's control actions (manipulation controller).}
    \label{fig:demo2}
\end{figure}

To introduce new skills, we collect three short demos, in which we set the original heading direction as a reference, the controller steers the robot to the original direction.
\begin{itemize}
    \item The robot is pushed over twice. It recovers from failure in left/right respectively, and steers to correct the heading direction.
    \item The robot is perturbed (to different heading direction) many times when moving forward. It steers to correct the heading direction after each perturbation.
    \item The robot goes over several obstacle with different size and shape using only one side of the track.
    
\end{itemize}

\begin{figure*}[t]
\centering
    \subfigure[]{
    \begin{minipage}[t]{0.14\linewidth}
    \centering
    \includegraphics[width=1\linewidth]{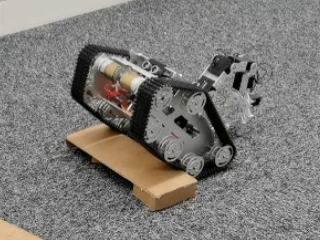}
    \end{minipage}%
    }
    \subfigure[]{
    \begin{minipage}[t]{0.14\linewidth}
    \centering
    \includegraphics[width=1\linewidth]{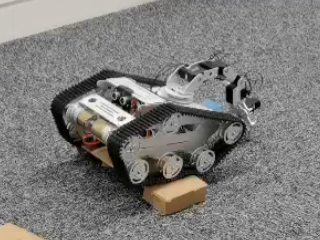}
    \end{minipage}
    }
    \subfigure[]{
    \begin{minipage}[t]{0.14\linewidth}
    \centering
    \includegraphics[width=1\linewidth]{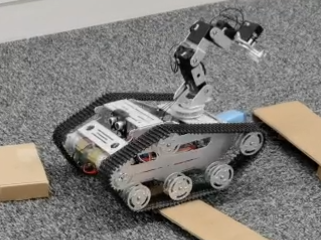}
    \end{minipage}
    }
    \subfigure[]{
    \begin{minipage}[t]{0.14\linewidth}
    \centering
    \includegraphics[width=1\linewidth]{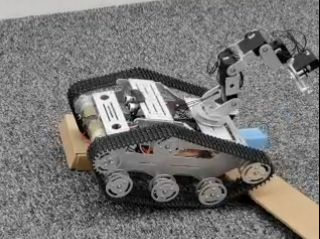}
    \end{minipage}
    }
    \subfigure[]{
    \begin{minipage}[t]{0.14\linewidth}
    \centering
    \includegraphics[width=1\linewidth]{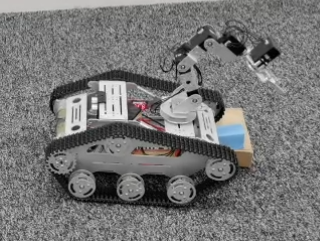}
    \end{minipage}
    }
    \subfigure[]{
    \begin{minipage}[t]{0.14\linewidth}
    \centering
    \includegraphics[width=1\linewidth]{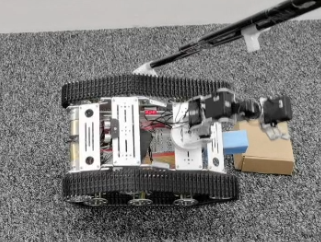}
    \end{minipage}
    }
    \subfigure[]{
    \begin{minipage}[t]{0.14\linewidth}
    \centering
    \includegraphics[width=1\linewidth]{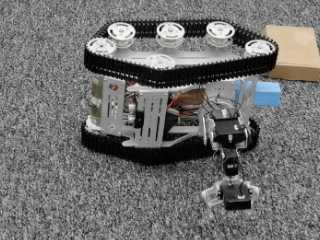}
    \end{minipage}
    }
    \subfigure[]{
    \begin{minipage}[t]{0.14\linewidth}
    \centering
    \includegraphics[width=1\linewidth]{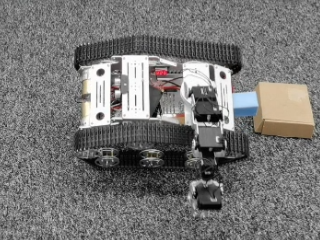}
    \end{minipage}
    }
    \subfigure[]{
    \begin{minipage}[t]{0.14\linewidth}
    \centering
    \includegraphics[width=1\linewidth]{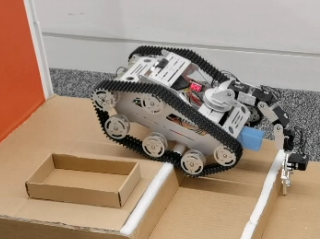}
    \end{minipage}
    }
    \subfigure[]{
    \begin{minipage}[t]{0.14\linewidth}
    \centering
    \includegraphics[width=1\linewidth]{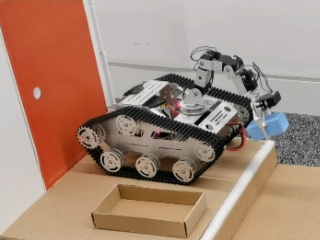}
    \end{minipage}
    }
    \subfigure[]{
    \begin{minipage}[t]{0.14\linewidth}
    \centering
    \includegraphics[width=1\linewidth]{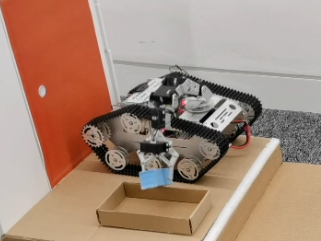}
    \end{minipage}
    }
    \subfigure[]{
    \begin{minipage}[t]{0.14\linewidth}
    \centering
    \includegraphics[width=1\linewidth]{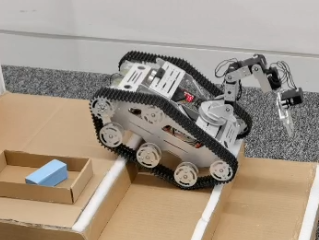}
    \end{minipage}
    }
    \caption{Test scenarios of the learned controller to achieve autonomous driving for obstacle negotiation, steering control, fall recovery and staircase ascent/descent. More extensive tests on uneven terrains can be found in the paper's accompanying video. }
    \label{fig:testsnapshot}
\end{figure*} 

\begin{figure}[t]
    \centering
    \includegraphics[width=0.5\textwidth]{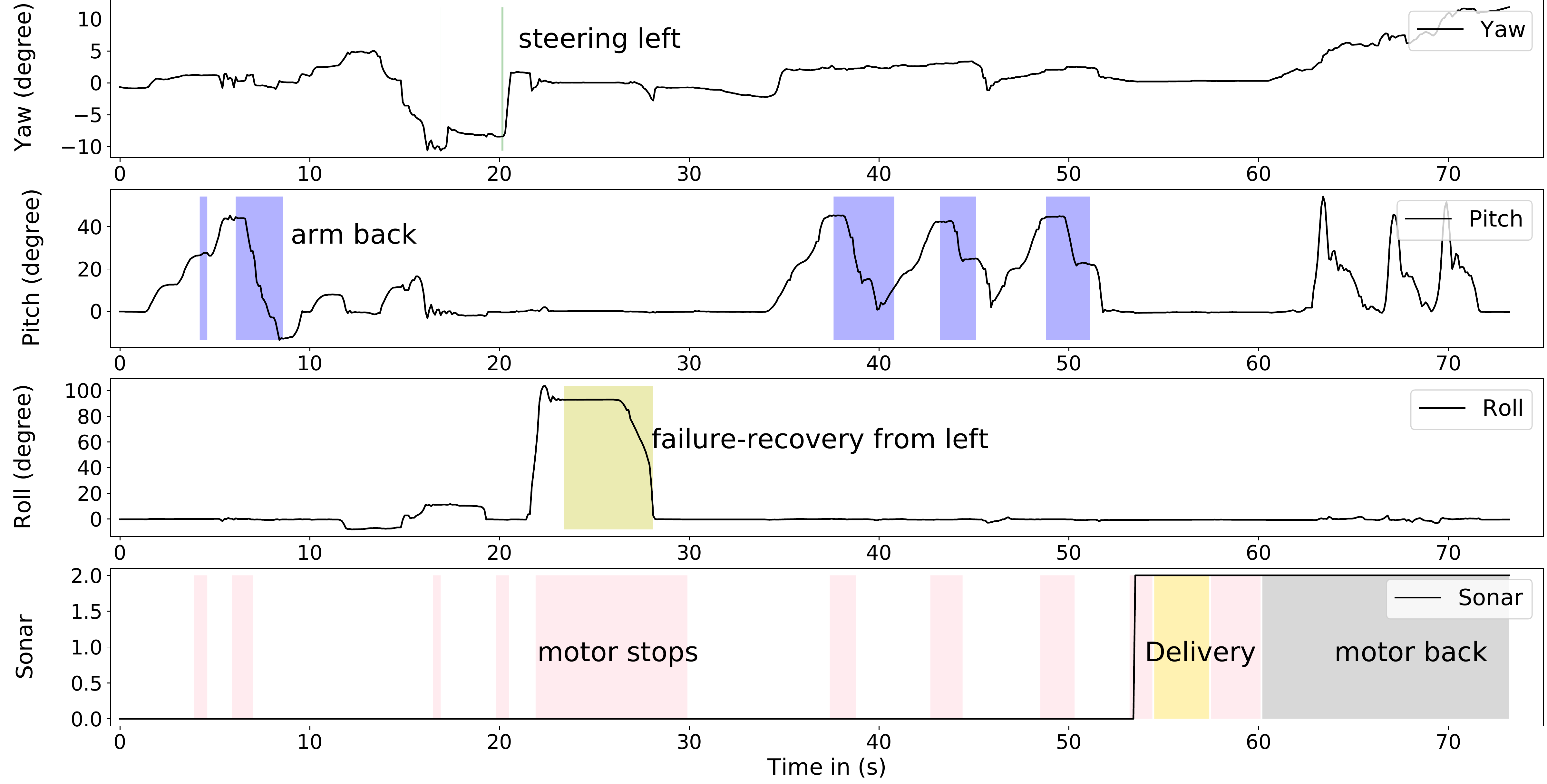}
    \caption{Observations and network's output actions in a real-world experiment.}
    \label{fig:test}
\end{figure}

By trimming and merging these 3 short demos together with the first demonstration, we obtain a long combined and near-optimal demo (the robot's yaw is $0\degree$ at the beginning): 

\begin{itemize}
    \item[(1)] The robot drives over 3 small obstacle. 
    \item[(2)] The robot was pushed over to left, then the expert triggers the failure-recovery-from-left action and steers right  to correct the heading direction.
    \item[(3)] The robot was toppled to the right, then the expert triggers the failure-recovery-from-right action and steers left.
    \item[(4)] The robot was perturbed to change the heading angle, then the expert steers to correct the yaw.
    \item[(5)] The robot drives over a large obstacle (with the help of arm) and a small obstacle.
    \item[(6)] The robot was toppled to the left, then the expert triggers the failure-recovery-from-left action and steers right.
    \item[(7)] The robot climbs a three-step stair with the help of arm.
    \item[(8)] The robot stops for 2 seconds when facing a wall (\textless 15 cm).
    \item[(9)] The robot goes backwards.
\end{itemize}

We use this new demonstration to train the reactive responsive control policy of the mobility controller. Fig.\ref{fig:demo1} shows the raw sensor data (black curves) of the merged demonstration, we mark expert's control actions with different colour shades. Fig.\ref{fig:demosnapshot} (1-9) show the actions in the demo.

\subsubsection{Manipulation controller}
The human expert teleoperates the robot to deliver the package when the robot is facing the wall (door), this demo includes:
\begin{itemize}
    \item[(1)] The robot goes forward to approach the wall. It stops to deliver the package.
    \item[(2)] The robot arm moves back to middle position.
\end{itemize}

Fig. \ref{fig:demo2} shows the raw sensor data (black curves) of the manipulation demo, we mark different actions with different colour shades. Fig.\ref{fig:demosnapshot} (10-12) show the delivery action in the demo.

\subsection{Results analysis and evaluation}
As there is only one demonstration for each controller, we do not have validation sets and test sets. The manipulation controller is easy to train, it converges after 800 steps and reach 95.37\% accuracy. The training process of mobility controller terminates at the 3500th step. The accuracy reaches 93.58\%. Our experiments and data indicate there is no over-fitting if the demo consists of enough amount of different robot observations and corresponding controls in different scenarios.

For behavioural cloning and LfD problems, training accuracy usually could not present the true performance of the algorithm. We change the size, position and yaw of the obstacles in the experimental arena setups to test the robustness of the controller.
The mobility controller could successfully handle these different setups.
Fig.\ref{fig:test} shows the robot observations and the control actions in real experiment and actions from the trained network controller. It can be inferred that our reactive responsive controller does not over-fit to the order of obstacles and the timing for each action. The control actions are based on a sequence of sensor inputs and the robot's global state. This method is very time-efficient and we think it is not difficult to be used by other behavioural cloning algorithms as long as the robot control interface is intuitive to human.

\subsection{Learning New Skills Efficiently Using Transfer Learning}
\label{subsec:TL}
Our robot system could learn new skills from a new edited demonstration. However, training the network from scratch is time-consuming. We use TL to greatly improve the learning efficiency to acquire new skills.

In our research, different control skills are generated from the same robot system using the same control interface. The Markov process of these control actions are similar and related. As a result, we restore the hyper-parameters of the three LSTM cells from the well-trained model in \ref{subsec:training}, and only initialise the fully-connected layer (red shade in Fig.\ref{fig:lstm}) while re-training. 
TL saves at least 60\% of training time in our experiments. Mobility controller network converges at around 1500 steps.


\section{Conclusion and Future Work}
\label{sec:conclusion}

This paper presents an LSTM-based control algorithm that enables an arm-equipped tracked robot to learn motion skills (self-steering, obstacle-negotiation, failure-recovery, stair-climbing and delivering) from teleoperated human demonstrations. We find that feeding a set of sequential data  containing several previous data and the current one to the LSTM controller would significantly improve its performance comparing to simply feeding the current data.  We also propose a method to minimise the effect of non-optimal demos for supervise-learning-based control algorithm and use transfer learning to speed up the training procedure. We decompose the delivery task into mobility part and manipulation part, and use 2 networks to control them separately. 

As for future work, we can introduce a path planner in the `High-level controller' shown in Fig.\ref{fig:controlDiagram}, which will enable more complex tasks.
In the training of aforementioned process, the control of motors and arm are decoupled. The human expert stops motor control while controlling the robot arm and vice-versa. We plan to learn whole-body control of arm-equipped mobile robot from several independent demos, or from an edited demonstration which contains those demonstrations.

\section{Acknowledgement}
This work has been supported by EPSRC UK Robotics and Artificial Intelligence Hub for Offshore Energy Asset Integrity Management (EP/R026173/1).

\bibliographystyle{IEEEtran}
\bibliography{IEEEabrv,IEEEexample}
\balance

\end{document}